\documentclass[nonacm]{acmart}
\AtBeginDocument{%
  }

\setcopyright{none}
\acmDOI{XXXXXXX.XXXXXXX}

\acmJournal{tweb}
\acmMonth{4}

\usepackage{latexsym}
\usepackage{enumitem}
\usepackage{multirow} 
\usepackage{array} 
\usepackage{longtable} 
\usepackage[T1]{fontenc}
\usepackage[utf8]{inputenc}
\usepackage{microtype}
\usepackage{subcaption} 
\usepackage{multicol}
\usepackage{colortbl}
\usepackage{xcolor}

\usepackage{subcaption} 
\usepackage{tcolorbox} 

\usepackage[main=english,bidi=default]{babel}

\babelprovide[import]{hindi}
\babelprovide[import]{sanskrit}

\newcommand{\hatexplain}{\textsc{HateXplain}}
\newcommand{\implicithate}{\textsc{ImplicitHate}}
\newcommand{\indohatemix}{\textsc{IndoHateMix}}

\begin{document}

\title{Rethinking Hate Speech Detection on Social Media: Can LLMs Replace Traditional Models?}

\author{Daman Deep Singh}
\authornote{Both authors contributed equally to this research.}
\affiliation{%
  \institution{Indian Institute of Technology Delhi}
  \country{India}
}

\author{Ramanuj Bhattacharjee}
\authornotemark[1]
\affiliation{%
  \institution{Indian Institute of Technology Kharagpur}
  \country{India}
}

\author{Abhijnan Chakraborty}
\affiliation{%
  \institution{Indian Institute of Technology Kharagpur}
  \country{India}
}

\renewcommand{\shortauthors}{Singh et al.}



\begin{abstract}
Hate speech detection across contemporary social media presents unique challenges due to linguistic diversity and the informal nature of online discourse. These challenges are further amplified in settings involving code-mixing, transliteration, and culturally nuanced expressions. While fine-tuned transformer models, such as BERT, have become standard for this task, we argue that recent large language models (LLMs) not only surpass them but also redefine the landscape of hate speech detection more broadly. To support this claim, we introduce \textsc{IndoHateMix}, a diverse, high-quality dataset capturing Hindi–English code-mixing and transliteration in the Indian context, providing a realistic benchmark to evaluate model robustness in complex multilingual scenarios where existing NLP methods often struggle. Our extensive experiments show that cutting-edge LLMs (such as, LLaMA-3.1) consistently outperform task-specific BERT-based models, even when fine-tuned on significantly less data. With their superior generalization and adaptability, LLMs offer a transformative approach to mitigating online hate in diverse environments. This raises the question of whether future works should prioritize developing specialized models or focus on curating richer and more varied datasets to further enhance the  effectiveness of LLMs.

\noindent
\textit{Disclaimer}: This paper contains content that readers may find offensive; unfortunately, it's an inherent aspect of studying hate speech.
\end{abstract}



\maketitle

\section{Introduction}
\label{sec:introduction}
Social media has transformed the way people interact, with various forms of interactions transcending geographical and linguistic barriers. However, this increased connectivity has also brought challenges, including the spread of hate speech, which can divide societies and harm vulnerable groups. 
\textit{Hate Speech} refers to any form of communication that expresses hostility, discrimination, or incites harm against an individual or group based on characteristics such as race, religion, ethnicity, gender, or political beliefs~\cite{rw1,rw2}. Hate speech is increasingly becoming a concern in many parts of the world. Its rapid spread over social media not only deepens social divisions but also threatens peace and harmony between communities~\cite{rw3}. Real-world incidents have starkly highlighted the dangers of unchecked hate speech. It has been a driving force behind major events such as the Rohingya crisis \cite{facebook-rohingya}, the $2020$ Delhi riots \cite{delhi-riots}, and the $2021$ U.S. Capitol riot \cite{capitol-riot}. Furthermore, incidents like the atrocities against minorities in Bangladesh~\cite{bangladesh-hatespeech} and ethnic violence in Ethiopia underscore how digital platforms can amplify extremism and fuel social unrest \cite{nzau2023confronting}.

Hate speech detection and mitigation is a critical task in Natural Language Processing (NLP). Much of the existing work has focused on fine-tuning transformer models like BERT and its variants, which have demonstrated strong performance in hate speech classification on certain datasets \cite{mbert,rw8}. However, these approaches are primarily limited to English-language contexts, restricting their applicability in multilingual settings. Prior to BERT, traditional machine learning techniques (e.g., SVM, Naive Bayes) and earlier neural network architectures used for hate speech detection also struggled to capture complex linguistic patterns. In countries like India, where \textit{code-mixing} — the blending of multiple languages in a single sentence — and \textit{transliteration} — writing one language using the script of another — are common, making the detection of hate speech in such context even more challenging \cite{intro1,intro2,intro3}. For instance, code-mixed languages like Hindi-English (Hinglish) and transliterated text introduce additional complexities due to informal syntax, variable grammar, and a lack of standardized datasets.


The emergence of large language models (LLMs) such as GPT-4o and LLaMA-3 has revolutionized NLP, demonstrating remarkable zero-shot and few-shot learning capabilities \cite{rw10,rw12}. Pre-trained on extensive and diverse corpora, these models require minimal task-specific fine-tuning and can adapt to new tasks with only a small set of high-quality samples~\cite{quality-data-llm, quality-data-llm-2}. This adaptability makes them highly effective at detecting both overt and subtle hate speech. However, the challenge intensifies in code-mixed and transliterated speech, where the blending of languages demands handling complex syntactic and semantic nuances \cite{rw2}. 
In this work, we aim to explore \textit{whether LLMs, with their broader linguistic adaptability and context awareness, surpass fine-tuned transformers in detecting hate speech} across diverse hate speech instances, including the complex multilingual environments that involve code-mixing and transliteration. Our investigation also questions \textit{whether developing specialized hate speech detection models for this task remains a worthwhile research endeavor.}

To this end, we conducted a comprehensive benchmarking study on three diverse datasets: \hatexplain\cite{hateXplain}, \implicithate\cite{implicitHate}, and our newly introduced \indohatemix. While \hatexplain\ and \implicithate\ include challenges like implicit vs. explicit hate speech classification, rationale-based explanations, and span-based toxicity annotations, they and other existing benchmarks largely overlook the complexities of code-mixed hate speech. To bridge this gap, we introduce \textsc{IndoHateMix}, specifically designed to tackle Hindi-English code-mixed hate speech -- a critical but underexplored research area. 

Through extensive experiments, we demonstrate that, indeed, LLMs significantly outperform traditional BERT-based transformers in hate speech detection. 
Our evaluation further demonstrates that recent open-source LLMs, like \texttt{LLaMA-3.1}, even when fine-tuned on just $10\%$ of the available data, outperform fine-tuned BERT-based models trained on the full dataset. The results highlight not only the superior generalization of LLMs in resource-constrained settings but also the significance of \indohatemix\ in appropriately benchmarking hate speech detection models in complex scenarios involving code-mixed and transliterated texts.
We complemented our benchmarking with in-depth error analysis and target-wise analyses to uncover where models stumble, and to distinguish whether hate is aimed at communities or individuals. These insights also underscore the value of a nuanced dataset like \indohatemix, which captures the subtle dynamics of real-world hate speech.



To summarize, our main contributions in this paper are as follows:
\begin{itemize}
    \item We conduct a comprehensive comparative analysis of state-of-the-art hate speech classification models -- from fine-tuned transformers to LLMs -- across multiple datasets that span diverse cultural and linguistic contexts.
    \item We introduce \indohatemix, a high-quality benchmark capturing Hindi–English code-mixing and transliteration, designed to reflect the linguistic and cultural nuances of the Indian online discourse.
    \item We provide empirical evidence showing that LLMs performs very well in this task, generalizing across different languages and cultural contexts, even in low-resource settings. This is supported by detailed error analysis, which reveals traditional models' limitations in hate speech detection in various scenarios.
\end{itemize}

\noindent
Our findings raise the question of whether future research in this area should prioritize developing specialized models or focus on curating more diverse datasets to further enhance LLMs' effectiveness in hate speech detection, contributing to safer and inclusive digital spaces.
\section{Related Works}
\label{sec:related_workx}
Considering its importance in maintaining the sanctity of online social media, research on Hate Speech has taken different routes -- some works investigated its impact (and patterns of attack) on individuals and groups, and others examined the detection and mitigation approaches. In this section, we attempt to cover research works belonging to both these categories. 

\subsection{Hate Speech: Patterns, Targets and Effects}
Many studies have explored the impact of hate speech on various communities, highlighting its pervasive presence both online and in offline settings. Hate speech often disproportionately targets marginalized and vulnerable groups, including religious minorities, women, Dalits, immigrants, and LGBTQ+ individuals \cite{silva2016analyzing}. These attacks are not limited to personal insults or isolated incidents; rather, they contribute to a broader environment of hostility and exclusion. The consequences can extend beyond individual experiences and instill widespread fear, inflict lasting emotional and psychological stress, and normalize discriminatory attitudes within the society.

In addition to detecting hate speech, it is often necessary to identify who is targeted and how the hate speech is communicated. 
This is especially important in multilingual countries where people writing in different languages use different ways to express things. 
\citet{talat2016hateful} studied hate speech on Twitter and looked at which groups were being targeted and how personal opinions might affect the way people label hate. 
\citet{hateXplain} developed 
an approach and a dataset that not only says whether a post is hateful or not but also explains why and shows which group is being targeted. 
Researchers have also looked at how hate speech occurs in different countries. In India, it is usually aimed at religious groups and is influenced by politics and community tensions~\cite{rajan2021insta}.  In Myanmar, hatred against the Rohingya was spread on Facebook, leading to actual physical violence leading to mass migration~\cite{stevenson2018facebook}. In Germany, hate on the Internet directed at immigrants has been associated with an increase in hate crimes~\cite{muller2021fanning}. The same is also observed in the US, particularly during elections when hate speech tends to target racial minorities and immigrants~\cite{mathew2019spread}. Another significant thing researchers discovered is that hate speech is not always explicit. Humans tend to hide it using code words, sarcasm, memes, or slang \cite{magu2018determining}. Because of this, it becomes more difficult for automatic systems to detect it. Additionally, because online content spreads quickly, these hidden forms of hate speech can reach many people rapidly.

\subsection{Detection and Mitigation Approaches}
Hate speech detection has been an active field of research, with numerous datasets and models developed to address this challenge. Several studies have explored hate speech detection in English, often contributing benchmark datasets such as HateXplain~\cite{hateXplain}, OLID~\cite{rw2}, ImplicitHate~\cite{implicitHate}, ToxicSpans~\cite{toxicSpans}, and Stormfront~\cite{rw3}. Recent efforts have also focused on low-resource languages 
including code-mixed and transliterated text~\cite{paper1, paper2, paper3, paper4, paper5}. 
The progression of hate speech detection models can be categorized into the following three phases: 

\begin{enumerate}
    \item \textbf{Pre-BERT Approaches:} Initial solutions used traditional machine learning methods like Support Vector Machines (SVM) \cite{rw4}, Naïve Bayes classifiers \cite{rw5}, and Random Forests \cite{rw6}. Additionally, rule-based and lexicon-based methods were widely used, where systems relied on predefined word lists and linguistic rules to identify hate speech \cite{rw6, rw4}.  These relied on manually crafted features such as n-grams and sentiment lexicons but struggled with contextual understanding and generalization.

    \item \textbf{BERT-Based Models:} The advent of transformer-based models, particularly BERT \cite{mbert}, significantly improved text classification by capturing contextual relationships more effectively. Fine-tuned BERT models like HateBERT \cite{rw8} outperformed conventional methods but were mostly trained on monolingual data and weak against code-mixed and transliterated text.

    \item \textbf{Large Language Models (LLMs)}: More recently, LLMs such as GPT-4 \cite{rw10}, T5 \cite{rw11}, and LLaMA \cite{rw12} have enabled zero-shot and few-shot learning for hate speech detection \cite{tweb1}. While LLMs show promise in multilingual settings, studies have shown that they often underperform on code-mixed and transliterated tasks due to the lack of explicit pretraining on such data \cite{rw13}.
\end{enumerate}


\noindent 
Earlier studies evaluating LLMs, such as \texttt{GPT-3.5} in zero-shot settings \cite{guo2023investigation, paper6}, have highlighted limitations in reasoning, explanation, and handling target community information, largely due to the linguistic complexities of code-mixed text, including inconsistent grammar, transliteration, and mixed syntactic structures \cite{priyadharshini2020named}. 
However, the emergence of newer LLMs trained on more diverse datasets calls for a re-evaluation of their effectiveness in detecting code-mixed and transliterated hate speech. 
In this paper, we bridge this gap.

Finally, going beyond finding hate speech, some researchers have also looked at ways to reduce its effect. One method is known as \textit{counterspeech}; this is where one responds to hate with nice or useful words, rather than remaining silent or fighting back with more hate. Research indicates that effective counter-speech can prevent hate from spreading and even transform the way individuals think \cite{mathew2019thou}. People use different styles for this, such as sharing facts, being humorous, being empathetic, or simply directly pointing out the problem. There is also some new research that attempts to employ AI to automatically generate these responses \cite{chung2019conan}. This matters because detecting hate speech alone is insufficient; we must also develop more effective and thoughtful ways to respond to it.

\section{Datasets}
\label{sec:dataset}
In this paper, we use two popular hate speech datasets, both bringing different insights to the task of detecting hate speech. Although these datasets offer useful insights, they are mostly geared towards monolingual environments. In order to bridge this gap between multilingual and code-mixed hate speech detection, we curate an additional novel dataset \indohatemix\ that is specifically created towards such linguistic diversity. This section explains these datasets and how they relate to our work.

\subsection{\hatexplain Dataset}
The \hatexplain dataset~\cite{hateXplain} presents a benchmark to reduce bias and provide explainability in hate speech detection. The annotations consist of three different tasks: (i) \textit{classification}, categorizing a post in one of the following categories: \textit{hate speech}, \textit{offensive}, or \textit{normal}; (ii) \textit{hate target}, the group being victimized by the post; and (iii) \textit{rationale}, indicating why certain text portions determine the classification. For our binary classification experiments, we focus only on posts labeled as hate speech and normal, excluding offensive entries. We use $594$ samples labeled as hate speech and $782$ samples labeled as normal from the official test set.
We include \hatexplain in our study due to its focus on explainability, which allows us to analyze how models interpret hate speech and assess biases in classification.

\subsection{\implicithate Dataset}
The \implicithate Dataset \cite{implicitHate} is another corpus developed to identify and analyze hate speech, mainly focusing on whether implicit and explicit terms relating to hateful messages are utilized. The dataset has $21,480$ tweets obtained from prominent extremist groups in the United States, and each entry has been labeled meticulously as being \textit{implicit\_hate}, \textit{explicit\_hate}, or \textit{not\_hate}. Beyond the level of categorization, the labels also capture the deeper intent or implied meaning contained in the content. We restructured the dataset to make it a binary for the purpose of this study; we labeled \textit{not\_hate} as $0$ and \textit{implicit\_hate} and \textit{explicit\_hate} as $1$. The transformation captures the difference between hate and non-hate messages while making the task of classification relatively easier. After preprocessing, we split the dataset into training and testing sets using an $80$:$20$ ratio and ensuring that the original class distributions are preserved through stratified sampling. This dataset is particularly relevant as it helps assess model performance in detecting implicit hate speech, a more challenging and context-dependent form of toxicity often overlooked in traditional hate speech detection models.

\begin{table*}[t]
\small
    \centering
    \begin{tabular}{|l|c|c|p{2cm}|c|}
        \hline
        \textbf{Dataset} & \textbf{Total Samples} & \textbf{Hate Speech} & \textbf{Non-Hate} & \textbf{Avg. Length} \\
        \hline
        \hatexplain \cite{hateXplain} & 12375  & 5342 (43.17\%) & 7033 (56.83\%)  & 23.5 words \\
        \implicithate \cite{implicitHate} & 21480 & 8189 (38.12\%) & 13291 (61.9\%) & 16.81 words \\
        \indohatemix (Our Contribution) & 11725 & 4457 (38.01\%) & 7268 (61.99\%) & 22.20 words \\
        \hline
    \end{tabular}
    \caption{Statistics of different datasets used in this study.}
    \label{tab:dataset_statistics}
\end{table*}

\subsection{\indohatemix Dataset}
The \indohatemix dataset is a key contribution of this study, 
addressing critical gaps in existing hate speech resources by capturing the cultural and situational nuances of Indian social media -- dimensions often overlooked in other datasets. 
A distinguishing feature of \indohatemix is its focus on \textit{code-mixing} and \textit{transliteration}, both of which are prevalent in India’s multilingual online communities. 
The linguistic complexity, informal tone, and strong context dependence of such content pose significant challenges for automated detection systems. By incorporating these real-world complexities, \indohatemix aims to advance research in hate speech detection within linguistically diverse and socially nuanced environments.

\subsubsection{Dataset Source} 
The \indohatemix dataset was curated from Koo~\cite{koo}, a microblogging platform that gained immense popularity in India after it launched in March $2020$. Branding itself as `India's Twitter', Koo accommodated multiple Indian languages promoting regional communication. 
This makes Koo an indispensable tool for studying hate speech because it gives deeper insight into the socio-political climate and cultural nuances within India. Our data collection period was between December 2023 and January 2024.

\subsubsection{Dataset Curation}
We used a keyword-based search strategy to identify accounts likely to generate hate speech. Keywords included terms such as Israel, Palestine, Ukraine, Hamas, Khalistan, Pakistan, Trudeau, China, Election, and Protest, targeting both global and local socio-political contexts. 
This approach enabled the identification of initial hate accounts, which were manually reviewed and categorized based on their activity levels. To expand the dataset, we examined the followers of these accounts and iteratively conducted additional searches. Different data collection strategies were employed for high-activity and low-activity accounts. For each account, a curated set of selected posts, referred to as candidate Koos, was maintained. Specifically, for low-activity accounts, the most recent $300$ posts were selected and, in the case of high-activity accounts, $300$ posts were chosen at regular intervals, with the spacing determined individually for each account based on its posting frequency. 
Since high-activity accounts often publish similar Koos throughout the day, this approach ensured diversity in the dataset. To achieve this, $50$\% of the candidate Koos were randomly selected for sampling, with posts extracted from one half and comments from the remaining Koos. This process contributed to curating the raw dataset, which was then refined through preprocessing steps to enhance data quality and relevance. These steps included removing URLs and emojis to reduce noise and bias, converting all text to lowercase for uniformity, and filtering out duplicates or near-duplicates using cosine similarity. Additionally, hashtags were retained but stripped of the `\#' symbol, and only posts containing between $5$ and $55$ words were considered to maintain contextual and grammatical adequacy.

\subsubsection{Annotation Process}
Our dataset annotation process relied on manual (human) annotation by three expert annotators, ensuring greater contextual accuracy than simple keyword-based identification. This approach enabled a careful analysis of each post’s context and overall sentiment to determine whether it constituted hate speech or not. For instance, in one example labeled as hate speech, the annotators identified the following post:

\begin{quote}
    \textit{``a clever \underline{enemy} is much better than a stupid friend. despite knowing fact, party leaders hug \& felicitate him."}
\end{quote}

Here, the use of terms like \underline{``enemy"} and the context of criticizing political leaders indicated a negative sentiment directed at a specific group, aligning with our definition of hate speech. 

In contrast, an example labeled as non-hate speech reads:

\begin{quote}
    \textit{``yes, they must condemn this \underline{dastardly act} in a loud and clear voice."}
\end{quote}

This post employs assertive language but does not target any specific group or express hatred. The annotators identified that it called for condemnation rather than inciting hostility, demonstrating the nuanced interpretation required for accurate annotation. 

The prevalence of code-mixed language, combining Hindi and English, posed a significant challenge. Negative sentiment expressed in one language could be amplified by words in the other. For example, in one post marked as hate speech, the user wrote:

\begin{figure}[!h]
\centering
\includegraphics[width=0.85\linewidth]{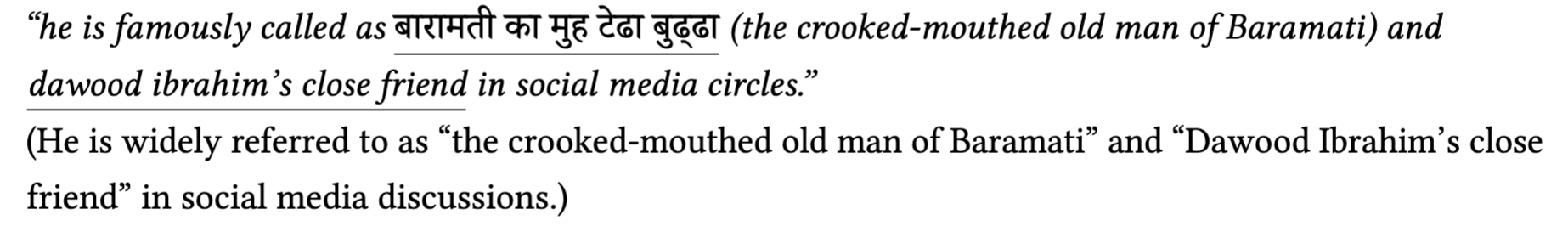}
\end{figure}

   

The Hindi phrase 
in the above example mocks the physical appearance and age of a political figure. Meanwhile, the English phrase links the person to a notorious criminal, reinforcing negative sentiment. This combination was categorized as hate speech due to its body-shaming and defamatory intent.

Conversely, a non-hate speech example reads:



\begin{figure}[!h]
\centering
\includegraphics[width=0.85\linewidth]{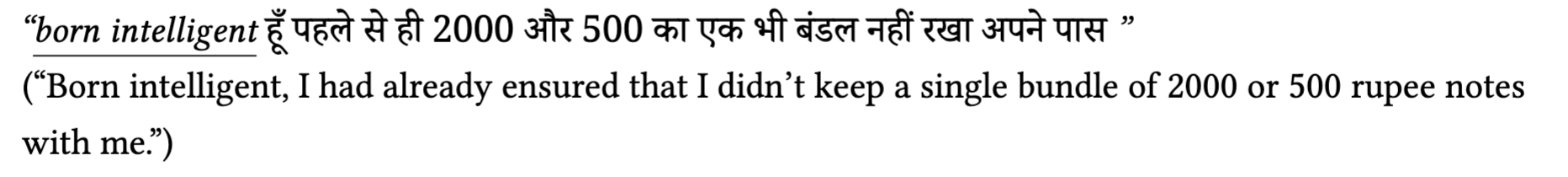}
\end{figure}

This post combines Hindi and English (``born intelligent'') in a light-hearted manner, joking about intelligence and not holding cash bundles. It was classified as non-hate speech because it lacked any malicious or harmful intent. 
Through this systematic approach, we navigated the linguistic complexities of Indian social media and accurately identified hate speech while maintaining cultural sensitivity. This process resulted in a high-quality dataset labeled with a binary scheme: \textbf{1 for hate speech and 0 for non-hate speech}. The dataset comprises a total of $11,725$ samples, with $4,457$ samples ($38.01$\%) labeled as hate and $7,268$ ($61.99$\%) categorized as non-hate.

To assess the reliability of our annotations and ensure the overall quality of the dataset, we computed inter-annotator agreement using the metric  Fleiss' Kappa~\cite{falotico2015fleiss}. 
We observed substantial agreement among annotators, with a Fleiss' Kappa of $0.685$. 
Following individual annotations, the annotators collaboratively reviewed ambiguous cases, reaching a consensus to ensure consistency and uniform application of standards across the dataset.

\vspace{1mm} \noindent
\textbf{Dataset statistics: } Table~\ref{tab:dataset_statistics} summarizes the number of samples, class distribution, and other relevant characteristics across different datasets, revealing 
the diversity of data sources and the representation of hate and non-hate instances.

\subsection{Targets of Hate}
Knowing who is targeted in hate speech allows us to understand the underlying social tensions and biases. 
Each of the datasets -- \indohatemix, \hatexplain, and \implicithate -- allows us to gain insight into the different types of groups or targets that are attacked most frequently.\footnote{\textit{Note that in the analysis, \textcolor{red}{red} is used to highlight religious communities (e.g., Hindus, Muslims, Jews), \textcolor{orange}{orange} for racial or ethnic groups (e.g., African Americans, Mexicans), and \textcolor{green}{green} for commonly targeted political figures.}}

In the \textbf{\indohatemix} dataset, hate speech incorporates religious as well as political targeting, demonstrating how individuals exhibit hostility not only against communities but also against individual leaders. Religious communities such as \colorbox{red!15}{\textcolor{red}{Hindus}}, \colorbox{red!15}{\textcolor{red}{Muslims}}, and \colorbox{red!15}{\textcolor{red}{Sikhs}} are repeatedly named with  hateful or violent language. For instance, \colorbox{red!15}{\textcolor{red}{Muslims}} are sometimes stereotyped as naturally aggressive or untrustworthy, and \colorbox{red!15}{\textcolor{red}{Sikhs}} are blamed for following separatist philosophies. These kinds of generalizations spread harmful stereotypes and deepen communal divides. At the same time, political hate is directed at leaders like \colorbox{green!15}{\textcolor{green!80!black}{Rahul Gandhi}}, \colorbox{green!15}{\textcolor{green!80!black}{Arvind Kejriwal}}, and \colorbox{green!15}{\textcolor{green!80!black}{Narendra Modi}}, all top leaders of major political parties in India. Unlike religious hate, these attacks focus on individuals and often reflect anger about governance, corruption, or other national issues. The tone is less about identity and more about disappointment or distrust toward these figures. 
This mix of religious and political hate is what makes \indohatemix unique. It reflects the actual nature of online hate in India, almost always written in code-mixed script (such as, Hindi-English), and can help us to develop systems that can address these types of nuanced local representations of hate. It illustrates that hate speech in India is as much about blaming individuals in authority as it is about targeting religious or caste groups.

In contrast, \textbf{\hatexplain} data is primarily concentrated on religious communities, particularly \colorbox{red!15}{\textcolor{red}{Muslims}} and \colorbox{red!15}{\textcolor{red}{Jews}}. The tone used here is mostly straightforward and aggressive, based on old stereotypes. For instance, \colorbox{red!15}{\textcolor{red}{Muslims}} are mentioned as violent or threatening, with \colorbox{red!15}{\textcolor{red}{Jews}} being blamed for exercising global power to serve hidden agendas. Very occasionally, \colorbox{orange!15}{\textcolor{orange!80!black}{Pakistanis}} are targeted as well, typically merging religious and national identity into a single category. What is striking in \hatexplain is the way hate is generalized: it does not target individuals, which is very often seen in \indohatemix, but rather holds entire communities accountable.

The \textbf{\implicithate} dataset looks at a different kind of hate -- one that is more hidden and suggestive. It tends to be directed towards racial and ethnic groups such as \colorbox{orange!15}{\textcolor{orange!80!black}{African Americans}}, \colorbox{orange!15}{\textcolor{orange!80!black}{Mexicans}}, and \colorbox{red!15}{\textcolor{red}{Muslims}}, but subtly so. Rather than employing insults, it employs negative stereotypes -- for example, that \colorbox{orange!15}{\textcolor{orange!80!black}{people with African heritage}} only manage to succeed in Western nations, or that \colorbox{orange!15}{\textcolor{orange!80!black}{Mexicans}} are uneducated. These remarks are less vitriolic on the surface but still propagate deleterious ideas.

These patterns in hate targets across different datasets are not just interesting facts -- they help us understand something deeper. By looking at who is being attacked in each dataset, we can see the social and political tensions behind the hate. Each dataset shows a different kind of hate: \textbf{\indohatemix} highlights direct political attacks, \textbf{\hatexplain} shows strong hate against religious and ethnic groups, and \textbf{\implicithate} reveals more hidden and subtle forms of bias. This tells us that hate detection models shouldn’t treat all data the same. Each dataset brings its own style and focus, so models must be trained and tested with these differences in mind. By comparing them this way, we can build better 
systems that handle hate more fairly and effectively.
\section{Experimental Evaluation}
As mentioned in the introduction, the primary goal of this paper is to benchmark the performance of various hate speech detection models. We attempt to cover large language models (LLMs) as well as traditional transformer models, such as BERT-based models, and custom APIs to determine which approach is more effective in detecting hate speech. Our analysis also includes analyzing how these models perform on various datasets, such as \indohatemix, \hatexplain, and \implicithate. To simulate real-world low-resource scenarios, we also conduct transfer learning experiments by fine-tuning models on limited subsets of data. Finally, we investigate the limitations of these models, particularly their handling of different cases of hate speech, through detailed error analysis.

\subsection{Models Selected for Benchmarking}
\label{subsec:models}
To ensure a comprehensive analysis, we benchmark a diverse range of models, which can be categorized into three groups: \textit{Multilingual BERT-based Models}, \textit{Hate Speech Detection APIs}, and \textit{Large Language Models}.

\subsubsection{Multilingual BERT-based Models}
We explore a group of prominent multilingual transformer-based models that have been pre-trained on multiple languages, including English and Hindi. Specifically, we evaluate the following models:
\begin{itemize}[noitemsep]
    \item \texttt{mBERT}: A multilingual BERT model proposed by~\citet{mbert}.
    \item \texttt{XLM-RoBERTa}: An extended multilingual RoBERTa model intoduced by~\citet{xlm-roberta}.
\end{itemize}




\subsubsection{Hate Speech Detection APIs}  
They offer off-the-shelf tools to identify harmful or offensive content without requiring any model training on the receiver's end. In our work, we use Google's Perspective API~\cite{perspectiveapi}, which returns six separate scores: \texttt{TOXICITY}, \texttt{SEVERE\_TOXICITY}, \texttt{IDENTITY\_ATTACK}, \texttt{INSULT}, \texttt{PROFANITY}, and \texttt{THREAT}\footnote{These are the evaluation categories provided by the Perspective API~\cite{perspectiveapi}.}. Because the API does not provide a single ``hate" score, we take the maximum of these six values as a proxy and classify content as hate speech whenever that maximum exceeds an arbitrary threshold of $0.5$.

\subsubsection{Large Language Models (LLMs)}  
Lastly, we explore a diverse set of state-of-the-art LLMs:

\begin{itemize}[noitemsep]
    \item \texttt{Mistral-7B}~\cite{mistral7b, mistral-7b-v0.3-hf}: An open-source, multilingual, efficient model featuring Sliding Window Attention for fast inference.
    \item \texttt{LLaMA3.1-8B}~\cite{llama3-models, llama-3.1-8b-hf}: An advanced open-source, long-context, multilingual model, with enhanced reasoning and safety via RLHF~\cite{rlhf}.
    \item \texttt{InternLM2.5-7B}~\cite{internlm2, internlm2_5-7b}: An efficiently trained, long-context, multilingual open-source LLM, with strong multilingual capabilities and enhanced safety alignment.
    \item \texttt{Qwen2.5-7B}~\cite{qwen2.5-7b}: A versatile open-source model designed for long context windows and robust multilingual performance.
    \item \texttt{GPT-4o-mini}~\cite{gpt-4o-mini}: A proprietary, highly optimized variant of GPT-4o~\cite{gpt-4o},  demonstrating enhanced reasoning performance and equipped with integrated safety mitigations.
\end{itemize}

\noindent
\textbf{Note.} All evaluated LLMs are open-source, with fully transparent architectures, known parameter counts, and the ability to be fine-tuned for our specific classification task, except for the proprietary \texttt{GPT-4o-mini}. Because it cannot be fine-tuned, we apply \texttt{GPT-4o-mini} zero-shot via the OpenAI API~\cite{gpt-4o-mini} using the same classification prompt as used for fine-tuning the open-source LLMs.

\subsection{Experimental Setup}
\label{sec:experiments}
We conducted all experiments on a single A$100$ $40$GB GPU. To optimize the efficiency of LLM-related experiments, we utilize the \texttt{llama-factory}~\cite{llama-factory} framework. For \hatexplain, we have retained the originally defined train-test-validation set; for \implicithate\ and \indohatemix\ datasets, we adopt a $70$-$20$-$10$ train-test-validation split.

\vspace{1mm} \noindent
\textbf{Prompt Template and Configuration: }
\label{subsec:impl_details}
In contrast to some prior studies~\cite{paper6, guo2023investigation}, we do not rely on complex prompt engineering techniques, such as Chain-of-Thought (CoT) reasoning~\cite{wei2023chainofthoughtpromptingelicitsreasoning} or explicitly specifying target categories within the prompt~\cite{paper6}, and instead we use a simple and direct prompting approach for LLMs. It is worth noting that 
different prompt engineering techniques can further boost LLM performance, and our findings establish a conservative baseline for LLM performance, and if it can outperform traditional models, further prompt engineering or tuning will only reinforce the assertion. 
In this study, all experiments utilize a consistent classification prompt and identical hyperparameter settings within each model family. Details are provided below:

\begin{itemize}[noitemsep, leftmargin=*]
  \item \textbf{BERT Fine-Tuning (mBERT \& XLM-RoBERTa):}
  \begin{multicols}{2}
  \begin{itemize}[noitemsep, topsep=0pt]
    \item Learning rate: $2e-5$
    \item Batch size: $2$
    \item Optimizer: AdamW (weight decay = $0.01$)
    \item Max sequence length: $128$
    \item Epochs: $3$
  \end{itemize}
  \end{multicols}

  \item \textbf{Perspective API:}  
  For each input, scores are computed for six categories: \texttt{TOXICITY}, \texttt{SEVERE\_TOXICITY}, \texttt{IDENTITY\_ATTACK}, \texttt{INSULT}, \texttt{PROFANITY}, and \texttt{THREAT}.  
  An input is labeled `Hate' if the maximum among all the above scores is greater than the threshold of $0.5$. \\


\item \textbf{LLMs (Mistral-7B, LLaMA3.1-8B, InternLM2.5-7B, Qwen2.5-7B, GPT-4o-mini):} 
\begin{itemize}[noitemsep, topsep=0pt]
    \item \textbf{Prompt Template:}  
    The following prompt was used across all LLM models:
    \begin{tcolorbox}[colback=gray!10, colframe=gray!50, title=Hate Speech Classification Prompt]
      \textbf{Instruction:} ``Review the provided statement and classify it as `Hate’ or `Non-Hate.’ Return `1’ for Hate and `0’ for Non-Hate.''\\
      \textbf{Input:} ``\texttt{\{statement\}}''\\
      \textbf{Output:} ``\texttt{\{label\}}''
    \end{tcolorbox}

    \item \textbf{Training and Testing Configuration:}
    \begin{multicols}{2}
    \begin{itemize}[noitemsep, topsep=0pt]
        \item Learning Rate: $5e$-$5$
        \item Batch Size: $1$
        \item Epochs: $3$
        \item Optimizer: AdamW
        \item LoRA Rank: $8$
        \item LoRA Alpha: $16$
        \item Temperature: $0.1$
        \item Top-p: $0.95$
        \item Max input length: $1024$ tokens
        \item Max output length: $256$ tokens
        \item Gradient accumulation steps: $1$
    \end{itemize}
    \end{multicols}
\end{itemize}

\end{itemize}

\begin{table*}[t!]
\small
\centering
\begin{tabular}{|l|p{5cm}|c|c|c|c|}
\hline
\multirow{2}{*}{\textbf{Model Type}} & \multirow{2}{*}{\textbf{Model Name}} & \multicolumn{4}{c|}{\textbf{Evaluation Metrics}} \\
                                     &                                      & \textbf{Accuracy} & \textbf{Precision} & \textbf{Recall} & \textbf{F1-Score} \\
\hline
\multicolumn{6}{|c|}{\textbf{\hatexplain}} \\
\hline
\multirow{2}{*}{Multilingual BERT-based Models} 
                                         & mBERT \cite{mbert}                & 0.87 & 0.87 & 0.87 & 0.87 \\
                                         & XLM-RoBERTa \cite{xlm-roberta}    & 0.86 & 0.86 & 0.86 & 0.86 \\
\hline
Hate Speech Detection APIs                & \colorbox{red!15}{\textcolor{red!70!black}{Perspective}} \cite{perspectiveapi} & \colorbox{red!15}{\textcolor{red!70!black}{0.66}} & \colorbox{red!15}{\textcolor{red!70!black}{0.68}} & \colorbox{red!15}{\textcolor{red!70!black}{0.68}} & \colorbox{red!15}{\textcolor{red!70!black}{0.66}} \\
\hline
\multirow{5}{*}{Large Language Models}   
                                        & \colorbox{green!15}{\textcolor{green!50!black}{LLaMA3.1-8B}} \cite{llama-3.1-8b-hf} & \colorbox{green!15}{\textcolor{green!50!black}{0.91}} & \colorbox{green!15}{\textcolor{green!50!black}{0.91}} & \colorbox{green!15}{\textcolor{green!50!black}{0.91}} & \colorbox{green!15}{\textcolor{green!50!black}{0.91}} \\
                                         & Mistral-7B \cite{mistral-7b-v0.3-hf}    & \textit{0.91} & \textit{0.91} & \textit{0.90} & \textit{0.90} \\
                                         & Qwen2.5-7B \cite{qwen2.5-7b}            & 0.90 & 0.90 & 0.90 & 0.90 \\
                                         & InternLM2.5-7B \cite{internlm2_5-7b}    & 0.89 & 0.90 & 0.88 & 0.89 \\
                                         & GPT-4o-mini \cite{gpt-4o-mini}          & 0.72 & 0.79 & 0.75 & 0.72 \\
\hline
\multicolumn{6}{|c|}{\textbf{\implicithate}} \\
\hline
\multirow{2}{*}{Multilingual BERT-based Models} 
                                         & mBERT \cite{mbert}                & 0.76 & 0.76 & 0.76 & 0.76 \\
                                         & \colorbox{red!15}{\textcolor{red!70!black}{XLM-RoBERTa}} \cite{xlm-roberta}    & \colorbox{red!15}{\textcolor{red!70!black}{0.65}} & \colorbox{red!15}{\textcolor{red!70!black}{0.65}} & \colorbox{red!15}{\textcolor{red!70!black}{0.65}} & \colorbox{red!15}{\textcolor{red!70!black}{0.65}} \\
\hline
Hate Speech Detection APIs                & Perspective \cite{perspectiveapi} & 0.69 & 0.67 & 0.64 & 0.64 \\
\hline
\multirow{5}{*}{Large Language Models}   
                                        & \colorbox{green!15}{\textcolor{green!50!black}{LLaMA3.1-8B}} \cite{llama-3.1-8b-hf} & \colorbox{green!15}{\textcolor{green!50!black}{0.82}} & \colorbox{green!15}{\textcolor{green!50!black}{0.82}} & \colorbox{green!15}{\textcolor{green!50!black}{0.81}} & \colorbox{green!15}{\textcolor{green!50!black}{0.81}} \\
                                         & Mistral-7B \cite{mistral-7b-v0.3-hf}   & \textit{0.82} & \textit{0.81} & \textit{0.81} & \textit{0.81} \\
                                         & Qwen2.5-7B \cite{qwen2.5-7b}           & 0.82 & 0.81 & 0.80 & 0.80 \\
                                         & InternLM2.5-7B \cite{internlm2_5-7b}   & 0.81 & 0.81 & 0.79 & 0.80 \\
                                         & GPT-4o-mini \cite{gpt-4o-mini}         & 0.70 & 0.70 & 0.71 & 0.70 \\
\hline
\multicolumn{6}{|c|}{\textbf{\indohatemix}} \\
\hline
\multirow{2}{*}{Multilingual BERT-based Models} 
                                         & mBERT \cite{mbert}                & 0.73 & 0.78 & 0.77 & 0.77 \\
                                         & XLM-RoBERTa \cite{xlm-roberta}    & 0.74 & 0.74 & 0.74 & 0.33\\
\hline
Hate Speech Detection APIs                & \colorbox{red!15}{\textcolor{red!70!black}{Perspective}} \cite{perspectiveapi} & \colorbox{red!15}{\textcolor{red!70!black}{0.70}} & \colorbox{red!15}{\textcolor{red!70!black}{0.79}} & \colorbox{red!15}{\textcolor{red!70!black}{0.61}} &  \colorbox{red!15}{\textcolor{red!70!black}{0.59}} \\
\hline
\multirow{5}{*}{Large Language Models}   
                                         & \colorbox{green!15}{\textcolor{green!50!black}{LLaMA3.1-8B}} \cite{llama-3.1-8b-hf}  & \colorbox{green!15}{\textcolor{green!50!black}{0.83}} & \colorbox{green!15}{\textcolor{green!50!black}{0.82}} & \colorbox{green!15}{\textcolor{green!50!black}{0.81}} & \colorbox{green!15}{\textcolor{green!50!black}{0.82}} \\
                                         & Mistral-7B \cite{mistral-7b-v0.3-hf}    & 0.81 & 0.81 & 0.79 & 0.80 \\
                                         & InternLM2.5-7B \cite{internlm2_5-7b}    & \textit{0.82} & \textit{0.81} & \textit{0.79} & \textit{0.80} \\
                                         & Qwen2.5-7B \cite{qwen2.5-7b}            & 0.81 & 0.81 & 0.78 & 0.79 \\
                                         & GPT-4o-mini \cite{gpt-4o-mini}          & 0.76 & 0.76 & 0.71 & 0.72 \\
\hline
\end{tabular}
\caption{Performance comparison of various hate speech detection models on \hatexplain, \implicithate, and \indohatemix\ datasets. The best results are highlighted in \colorbox{green!15}{\textcolor{green!50!black}{green}} while the worst are in \colorbox{red!15}{\textcolor{red!70!black}{red}}. Across all three datasets, LLMs have maintained superior performances.}
\label{tab:benchmarking_all}
\end{table*}

\subsection{Benchmarking Results}
\label{subsec:results}
Next, we present the benchmarking performance of various models on the \hatexplain, \implicithate, and \indohatemix\ datasets, evaluated across accuracy, precision, recall, and F1-score. The results from benchmarking on these datasets is summarized in Table~\ref{tab:benchmarking_all}. 
As we can observe from the table, across all datasets, \texttt{LLaMA3.1-8B} consistently delivers the highest performance, showcasing its superior ability to generalize in diverse hate speech detection settings. Specifically, it outperforms the BERT-based models by nearly $4$\%, $8$\%, and $14$\% in terms of accuracy on the \hatexplain, \implicithate, and \indohatemix\ datasets, respectively. Other LLMs -- \texttt{InternLM2.5-7B} and \texttt{Mistral-7B} -- follow closely behind, significantly outperforming multilingual BERT-based models and the popular hate speech detection API (Perspective) across all datasets. Interestingly, open-source LLMs -- \texttt{LLaMA3.1-8B}, \texttt{Mistral-7B}, \texttt{InternLM-2.5-7B}, \texttt{Qwen2.5-7B} -- outperform the proprietary \texttt{GPT-4o-mini} model in the hate speech detection task. We emphasize that, in this comparison, GPT-4o-mini is not fine-tuned as its exact size and training details are undisclosed, which makes direct comparison tricky -- any gap may reflect the differences in scale or pretraining rather than model quality. Even if it matched or beat our open-source LLMs with fine-tuning~\cite{azure-gpt-finetune}, it would only reinforce our claim that modern LLMs, when properly adapted, outperform traditional approaches, while open models still offer the benefits of transparency and customization.

\subsection{Transfer Learning and Limited Data Experiments}
\begin{figure}[t]
  \centering
  \begin{subfigure}{0.47\textwidth}
    \centering
    \includegraphics[width=0.75\linewidth]{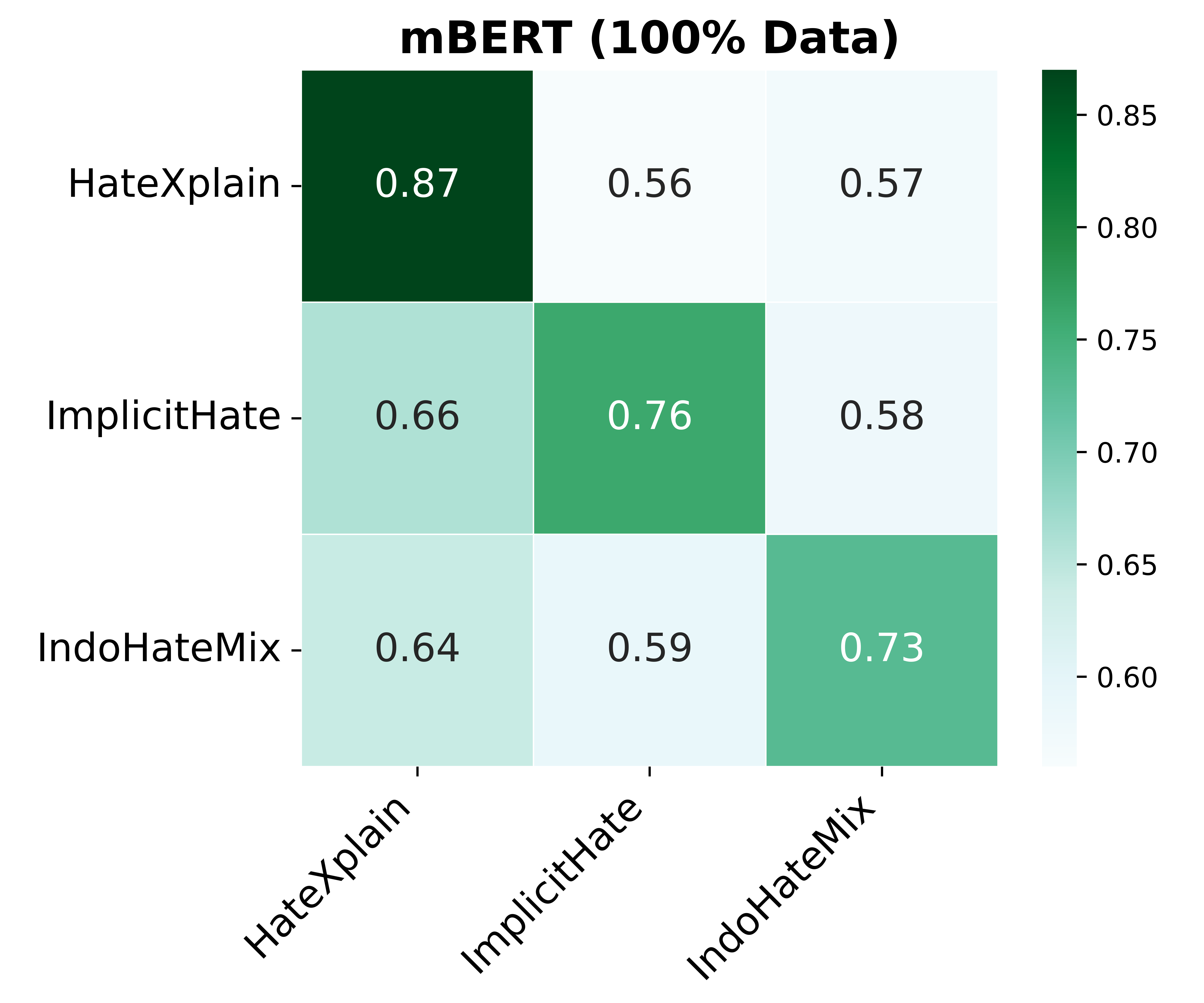}
    \caption{mBERT transfer learning (full data)}
    \label{fig:heatmap_mbert_full_contrast}
  \end{subfigure}
  \hfill
  \begin{subfigure}{0.47\textwidth}
    \centering
    \includegraphics[width=0.75\linewidth]{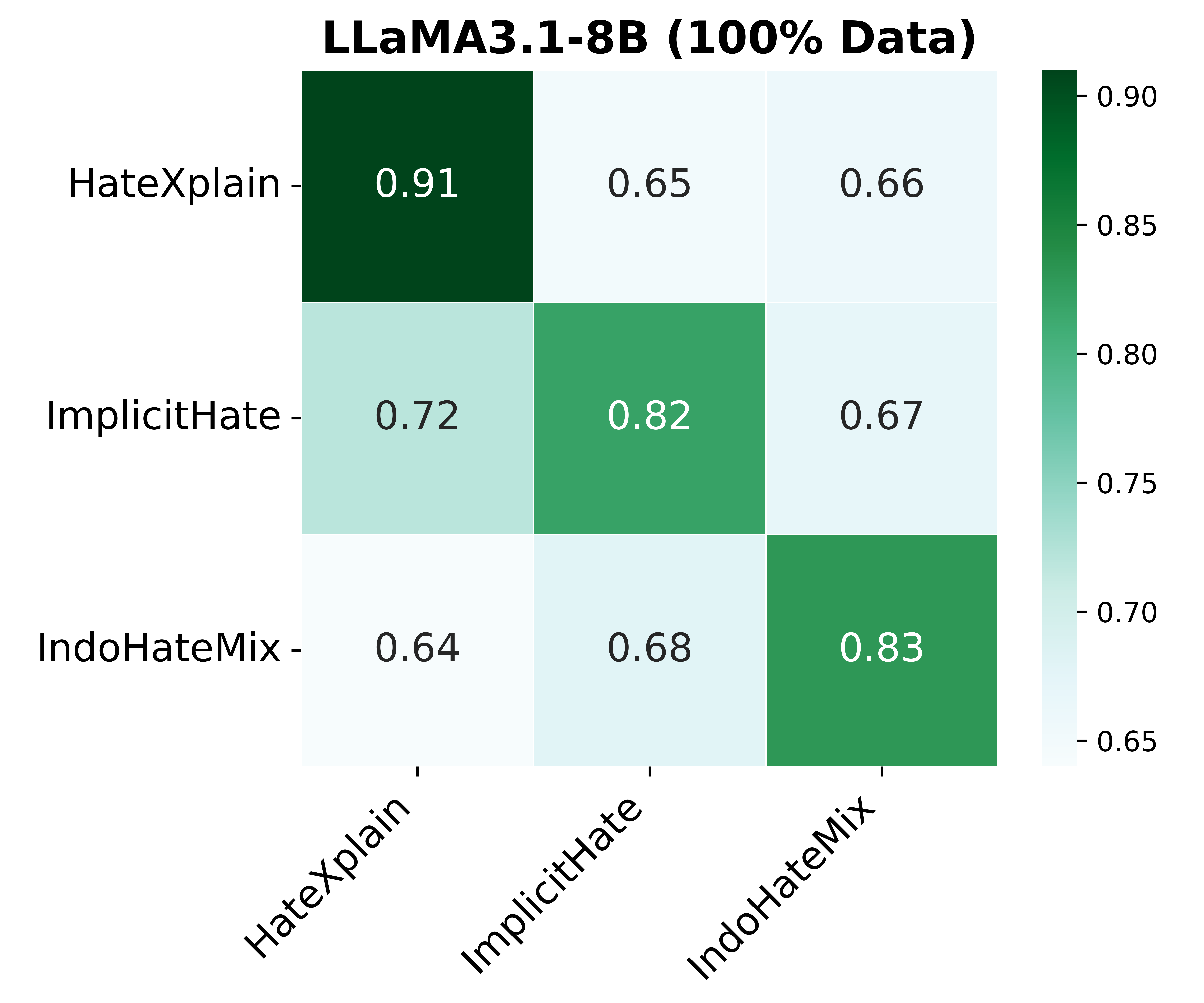}
    \caption{LLaMA3.1-8B transfer learning (full data)}
    \label{fig:heatmap_llm_full_contrast}
  \end{subfigure}
  \caption{Contrast heatmaps of cross-dataset transfer learning using full training data.}
  \label{fig:transfer_heatmaps_full}
\end{figure}

To evaluate the transfer learning capabilities of the state-of-the-art models for the hate speech detection task, we compare the best-performing LLM \texttt{LLaMA3.1-8B}, against the top BERT-based model \texttt{mBERT}, as identified in our earlier benchmarking experiments~\ref{tab:benchmarking_all}. Fig.~\ref{fig:transfer_heatmaps_full} summarizes the results of these transfer learning experiments, where models fine-tuned on one dataset are evaluated across all datasets. The rows indicate the datasets used for fine-tuning the corresponding model, while the columns represent the test datasets. Across all configurations, \texttt{LLaMA3.1-8B} consistently outperforms \texttt{mBERT}. For instance, when trained on the \hatexplain\ dataset and tested on all three datasets, the relative gain in accuracy of \texttt{LLaMA3.1-8B} over \texttt{mBERT}, averaged across the three test sets is a staggering $12.15$\%. This stark performance gap highlights \texttt{LLaMA3.1-8B}'s remarkable ability to generalize across diverse linguistic and cultural contexts, even when trained on limited domain-specific data.

These findings are particularly significant for real-world applications, where labeled data for niche challenges like code-mixing and transliteration is scarce. Unlike \texttt{mBERT}, which struggles to adapt to unseen linguistic patterns, \texttt{LLaMA3.1-8B} leverages its pretrained multilingual world knowledge to infer context and intent, even in low-resource settings. This adaptability makes \texttt{LLaMA3.1-8B} a powerful tool for combating hate speech in linguistically diverse regions, where traditional models often fall short.

\begin{figure}[!t]
  \centering
  \begin{subfigure}[t]{0.47\textwidth}
    \centering
    \includegraphics[width=0.75\linewidth]{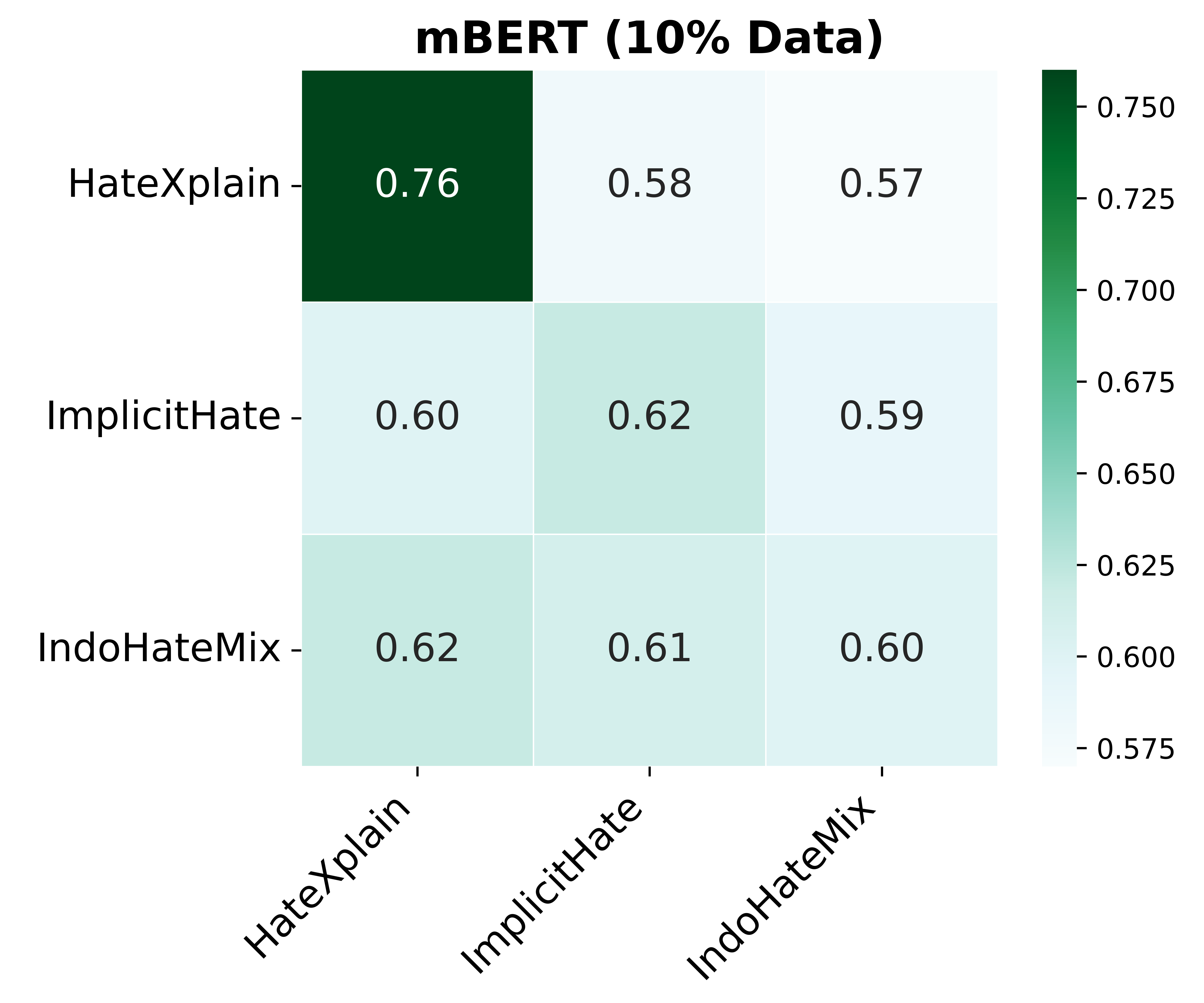}
    \caption{mBERT transfer learning (10\% data)}
    \label{fig:heatmap_mbert_10_contrast}
  \end{subfigure}
  \hfill
  \begin{subfigure}[t]{0.47\textwidth}
    \centering
    \includegraphics[width=0.75\linewidth]{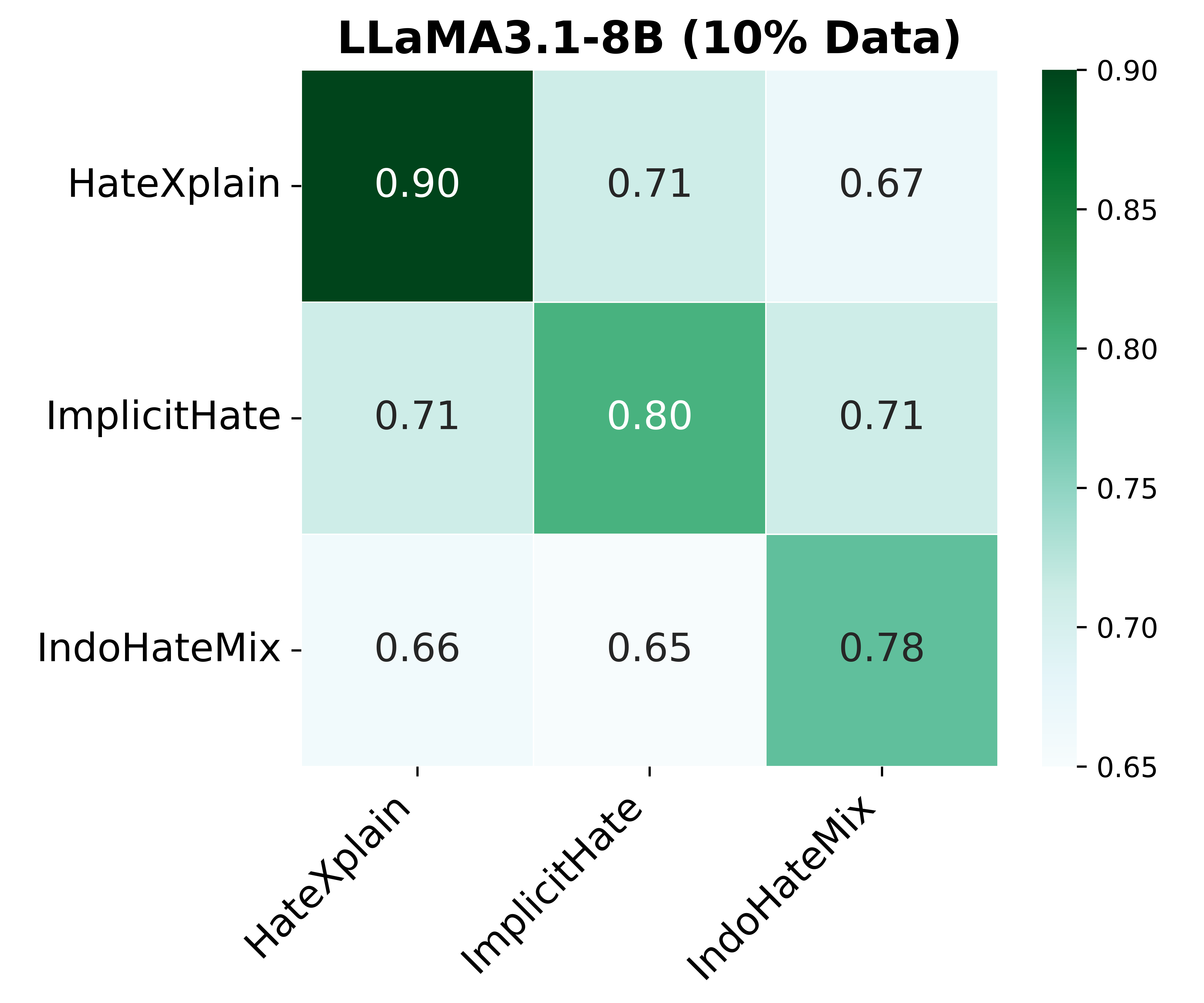}
    \caption{LLaMA3.1-8B transfer learning (10\% data)}
    \label{fig:heatmap_llm_10_contrast}
  \end{subfigure}
  \caption{Contrast heatmaps of transfer learning with limited (10\%) training data.}
  \label{fig:transfer_heatmaps_10}
\end{figure}

Further extending the transfer learning experiment, Fig.~\ref{fig:transfer_heatmaps_10} presents model performance in a \textit{limited data setting}, where only a random $10$\% (sampled without replacement) subset of the available data is used for fine-tuning\footnote{Note that the results presented in Figure~\ref{fig:transfer_heatmaps_10} are the average of $5$ experimental runs to ensure robustness and to reduce variance in the evaluation metrics.}. Despite the limited data, \texttt{LLaMA3.1-8B} consistently outperforms \texttt{mBERT} across all datasets, demonstrating its robustness even in data-constrained environments. 

Remarkably, even when fine-tuned on just $10$\% of the data using a rank-$8$ LoRA adapter that adds only about $4$ million trainable parameters~\cite{lora-paper} -- far fewer than \texttt{mBERT}’s $179$ million~\cite{mbert-hf} -- \texttt{LLaMA3.1-8B} still outperforms \texttt{mBERT} fine-tuned on the full dataset (see Table \ref{tab:benchmarking_all}), underscoring their efficiency and ability to generalize from minimal data. This finding strengthens our position that enhancing LLMs through curated, diverse datasets may offer greater returns than building task-specific models~\cite{quality-data-llm, summarization-dead}.

\subsection{Error Analysis}

\begin{figure*}[!t]
\centering
\includegraphics[width=\textwidth]{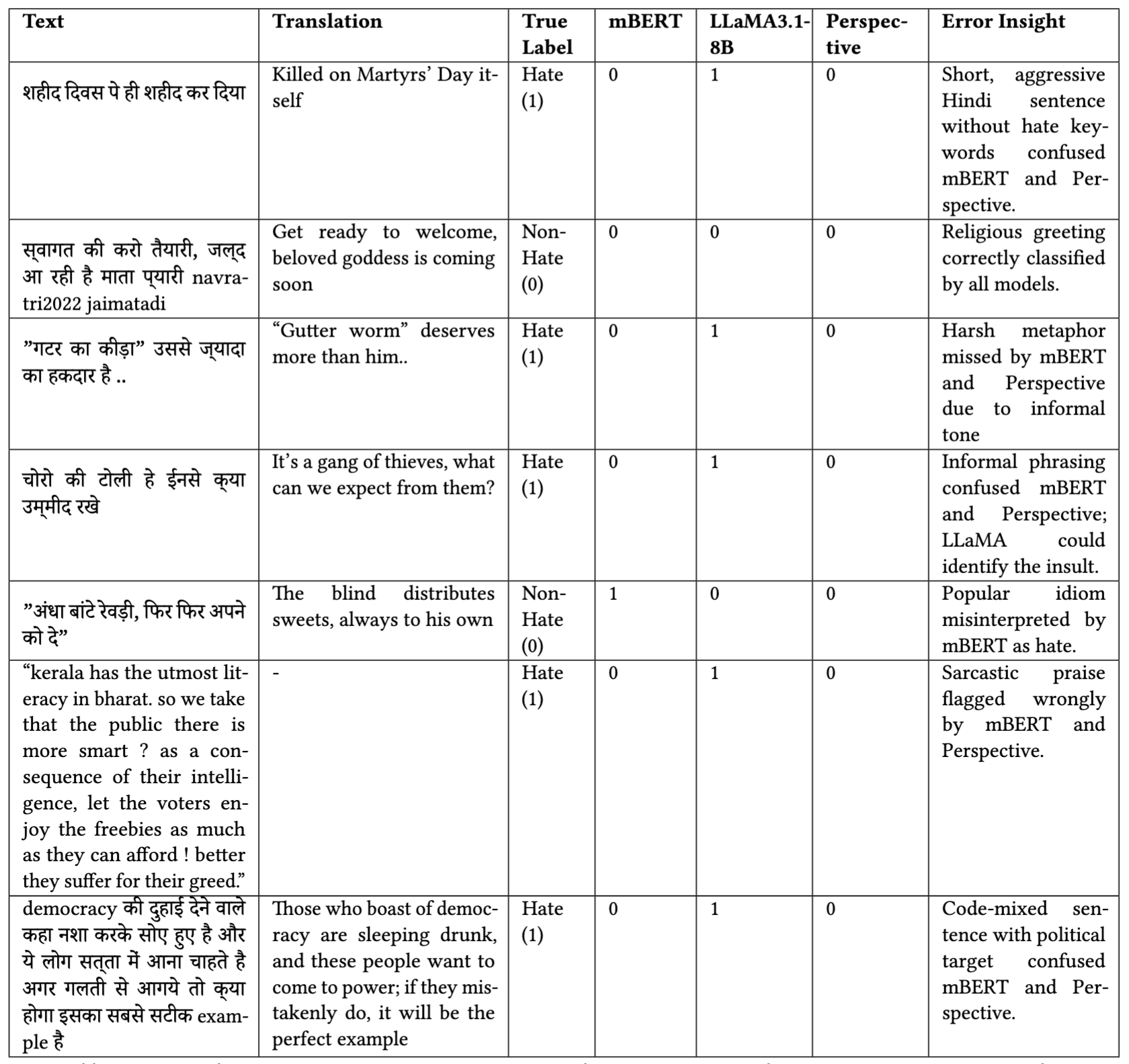}
\caption{Error analysis comparing mBERT, LLaMA3.1-8B, and Perspective API predictions on \indohatemix\ examples.}
\label{tab:error_analysis}
\end{figure*}

To gain a better insight into where the models are going wrong, we closely examined their mistakes. Of all the fine-tuned transformer models we experimented with, \texttt{mBERT} and \texttt{LLaMA3.1-8B} and Perspective API performed the best in their respective categories — with mBERT standing for conventional multilingual models and \texttt{LLaMA3.1-8B} standing for the new large language models (LLMs). Therefore, we selected these three for a detailed comparison. Table~\ref{tab:error_analysis} contains some exemplary cases from the \indohatemix dataset in which \texttt{mBERT}'s predictions deviate from those of \texttt{LLaMA3.1-8B}, with the former being wrong. Some such cases include a diverse set of scenarios, including exclusive Hindi, Hindi-English code-mixed posts, political satire, and informal conversational content.

From these examples, we observe clear differences in how the three models handle hate speech detection. \textbf{mBERT} frequently fails to identify hate speech when it is expressed in short, informal, or indirect forms (Examples 1, 3, and 4). It also struggles with longer, code-mixed sentences, such as in Example 7, where hateful content is embedded within political commentary. Furthermore, \texttt{mBERT} often misclassifies non-hateful content as hate speech and vice versa, especially in the presence of sarcasm or emotionally charged language (Examples 5 and 6). These cases suggest that \texttt{mBERT} is sensitive to surface-level cues like tone but often fails to grasp the deeper semantic intent.

In contrast, \textbf{LLaMA3.1-8B} demonstrates more robust performance. It successfully detects hate speech in challenging scenarios that \texttt{mBERT} misses (Examples 1, 3, 4, and 7), and it avoids false positives in posts involving sarcasm or humor (Examples 5 and 6). This indicates a stronger contextual and tonal understanding, allowing it to move beyond literal interpretation toward more nuanced comprehension. 
\textbf{Perspective API}, while useful in straightforward cases, appears more limited in scope. Like \texttt{mBERT}, it frequently fails to detect hate speech when the expression is indirect or code-mixed between English and Hindi (Examples 1, 3, 4, and 7). It also misclassifies sarcastic language, as seen in Example 6. These limitations highlight the API's difficulty in processing the informal, multilingual, and often ambiguous language typical of Indian social media platforms.

These misclassifications are significant, as they underscore the challenges of hate speech detection in linguistically diverse and informal online environments. mBERT, in particular, exhibits a tendency to flag benign content -- such as political opinions or jokes—as hate speech, while missing actual harmful content. In contrast, LLMs like LLaMA3.1-8B are more adaptable to the code-mixed and conversational tone of social media posts. Their flexibility allows them to more accurately identify hate speech without overgeneralizing.

\vspace{1mm} 
\noindent \textbf{Takeaway:} Overall, our analyses in this section highlight the comparative advantage of large language models (LLMs) in multilingual and informal contexts such as those found in India. These models generalize more effectively across varied styles of writing and better handle the noisy, real-world language seen online. This suggests a promising direction for future work: rather than developing multiple narrow, task-specific models, it may be more effective to invest in powerful, general-purpose LLMs trained on diverse, culturally grounded datasets to combat online hate more reliably.

\section{Conclusion}
\label{sec:conclusion}
In this work, we examined whether recent advances in large language models, noted for their strong linguistic competence and contextual understanding, can outperform traditional models that are and optimized for the critical task of hate speech detection. 
To this end, we introduced \textbf{\indohatemix}, a novel benchmark dataset specifically curated to address the unique linguistic and cultural challenges of the Indian social media landscape. With two more additional datasets, we conducted a comprehensive evaluation across a spectrum of models, ranging from widely used transformer-based baselines such as \texttt{mBERT}, to state-of-the-art LLMs like \texttt{LLaMA-3.1-8B}. Our experiments consistently showed that LLMs significantly outperform traditional models, even in low-resource scenarios where labeled data is scarce. Specifically, \texttt{LLaMA-3.1-8B} demonstrated superior ability to generalize across languages, detect indirect and sarcastic expressions of hate, and accurately interpret contextually rich or ambiguous posts -- areas where BERT based models showed considerable weaknesses.

These findings underscore the transformative potential of LLMs in improving hate speech detection systems. Their adaptability to handle informal, multilingual content position them as a promising direction for future work in this space. Rather than relying on specialized dataset-specific solutions, our results suggest that leveraging general-purpose LLMs, especially when fine-tuned with culturally and linguistically diverse datasets like \textbf{\indohatemix}, can lead to more reliable and context-sensitive hate speech detection frameworks.

\subsection{Limitations and Future Work}
While our study demonstrates superior performance of LLMs for hate speech detection, 
it is important to acknowledge several limitations that could influence the scope and robustness of our findings, along with possible directions for future work.

\noindent
\textbf{Dataset Specificity and Scope}: Our evaluation is conducted on three datasets: \hatexplain, \implicithate, and our newly introduced \indohatemix. While these datasets offer diverse challenges, including implicit hate speech and code-mixing, they may not fully represent the entire spectrum of online hate speech. Specifically, \indohatemix is curated from the Koo platform, focusing on Hindi-English code-mixed data which reflects the socio-political context of India. Since different platforms are known to have intrinsic biases~\cite{chakraborty2016dissemination, chakraborty2017makes}, one could expand the study to include multilingual and multimodal datasets from different platforms 
to evaluate generalizability across broader contexts.

\noindent
\textbf{Linguistic and Cultural Context}: The complexities of code-mixed languages, such as Hinglish, introduce challenges due to informal syntax, variable grammar, and transliteration. Although LLMs show improved adaptability, the models' understanding is still developing, especially concerning nuanced cultural references and implicit meanings. The annotation process, while thorough, relies on human annotators, introducing potential biases despite efforts to ensure consistency and mitigate bias. Further exploration can be done in the direction of culturally grounded embeddings or fine-tuning approaches that better capture regional sociolinguistic cues.

\noindent
\textbf{Model Biases}: Our analysis seeks to mitigate biases by using explainable datasets like \hatexplain to understand how models detect hate speech. However, bias can take different forms, and our study primarily focuses on classification accuracy and generalization across linguistic variations. A more comprehensive bias analysis involving demographic, gender, or socio-economic factors remains a promising avenue for future work.

\noindent
\textbf{Prompting and Fine-Tuning:} To ensure a fair comparison between traditional models and LLMs, we adopted a straightforward prompting strategy, deliberately avoiding advanced methods like Chain-of-Thought (CoT) reasoning. While CoT has been shown to significantly improve LLM performance on complex reasoning tasks~\cite{wei2023chainofthoughtpromptingelicitsreasoning}, our aim was to evaluate the inherent capabilities of LLMs with minimal task-specific tuning. Additionally, we did not explore recent advances in reasoning-driven interpretability techniques, which could identify hate speech targets and contribute to more robust and interpretable detection systems. Incorporating such techniques could be an important step for future work focused on robust and explainable hate speech detection.

\noindent
\textbf{Computational Resources and Model Size}: Our experiments were conducted on a single A$100$ $40$GB GPU, which may limit the scalability and exploration of larger models or more extensive hyperparameter tuning. While we benchmarked several state-of-the-art LLMs, including \texttt{LLaMA3.1-8B} and \texttt{GPT-4o-mini}, the rapidly evolving landscape of LLMs means that newer models with different architectures and training data may yield different results. Future work could investigate more efficient yet accurate hate speech classification using quantized versions of larger language models.

\noindent
\textbf{Generalization to Low-Resource Languages}: Although our study includes code-mixed data, the primary languages are Hindi and English. The effectiveness of LLMs in detecting hate speech in other low-resource languages or language combinations may vary due to differences in linguistic structures, available data, and cultural contexts. Exploring adaptation techniques and cross-lingual transfer learning for other underrepresented language pairs presents a valuable future direction.

\noindent
\textbf{Ethical Considerations}: Studying hate speech inherently involves dealing with offensive content. While we have taken precautions to minimize exposure and ensure responsible research practices, the potential for unintended consequences, such as reinforcing harmful stereotypes, remains a concern.

\vspace{1mm} \noindent
Finally, this work highlights the need to solve hate speech detection through the lens of general-purpose  large language models. As we continue to confront the toxic online discourse, such models offer a scalable and inclusive path forward, and we hope that this work can act as a torchbearer towards that road.






\bibliographystyle{ACM-Reference-Format}
\bibliography{main}












\end{document}